\documentclass[letterpaper, 10 pt, journal, twoside]{IEEEtran}
\usepackage{times}
\usepackage{graphicx}
\usepackage{subfigure}
\usepackage{amsmath}
\usepackage{amssymb}
\usepackage{amsfonts}
\usepackage{comment}
\usepackage{color}
\usepackage{multicol}
\usepackage{booktabs}
\usepackage{cite}
\usepackage{multirow}  
\usepackage{notoccite}
\usepackage[bookmarks=true]{hyperref}
\usepackage[font=small,labelfont=bf]{caption}
\usepackage[top=1in, left=0.75in, bottom=0.75in, right=0.75in]{geometry}
\usepackage{float}

\usepackage{pifont}

\IEEEoverridecommandlockouts
\begin{document}

\title{360FusionNeRF: Panoramic Neural Radiance Fields with Joint Guidance}

\markboth{IEEE Robotics and Automation Letters. Preprint Version. September 2022}
{Shreyas \MakeLowercase{\textit{et al.}}: 360FusionNeRF}

\author{Shreyas Kulkarni$^{1}$, Peng Yin$^{2,*}$, and Sebastian Scherer$^{2}$

\thanks{

$^{1}$S. Kulkarni is with the Department of Mechanical Engineering, Indian Institute of Technology, Madras {(shreyas.kulkarni@smail.iitm.ac.in)}.

$^{2}$P. Yin and S. Scherer are with Robotics Institute, Carnegie Mellon University, Pittsburgh, PA 15213, USA {(pyin2, basti@andrew.cmu.edu)}.}
}

\maketitle


\begin{abstract}
We present a method to synthesize novel views from a single $360^\circ$ panorama image based on the neural radiance field (NeRF). 
Prior studies in a similar setting rely on the neighborhood interpolation capability of multi-layer perceptions to complete missing regions caused by occlusion, which leads to artifacts in their predictions.
We propose 360FusionNeRF, a semi-supervised learning framework where we introduce geometric supervision and semantic consistency to guide the progressive training process. Firstly, the input image is re-projected to $360^\circ$ images and auxiliary depth maps are extracted at other camera positions. The depth supervision, in addition to the NeRF color guidance, improves the geometry of the synthesized views. 
Additionally, we introduce a semantic consistency loss that encourages realistic renderings of novel views.
We extract these semantic features using a pre-trained visual encoder such as CLIP, a Vision Transformer trained on hundreds of millions of diverse 2D photographs mined from the web with natural language supervision. 
Experiments indicate that our proposed method can produce plausible completions of unobserved regions while preserving the features of the scene. When trained across various scenes, 360FusionNeRF consistently achieves the state-of-the-art performance when transferring to synthetic Structured3D dataset (PSNR $\sim$ 5\%, SSIM $\sim$3\% LPIPS $\sim$13\%), real-world Matterport3D dataset (PSNR $\sim$3\%, SSIM $\sim$3\% LPIPS $\sim$9\%) and Replica360 dataset (PSNR $\sim$8\%, SSIM $\sim$2\% LPIPS $\sim$18\%). 
We provide the source code at
\href{https://github.com/MetaSLAM/360FusionNeRF}{https://github.com/MetaSLAM/360FusionNeRF}.
\end{abstract}



\section{Introduction}
\begin{figure*}[!ht]
    \centering
  \includegraphics[width=0.99\linewidth]{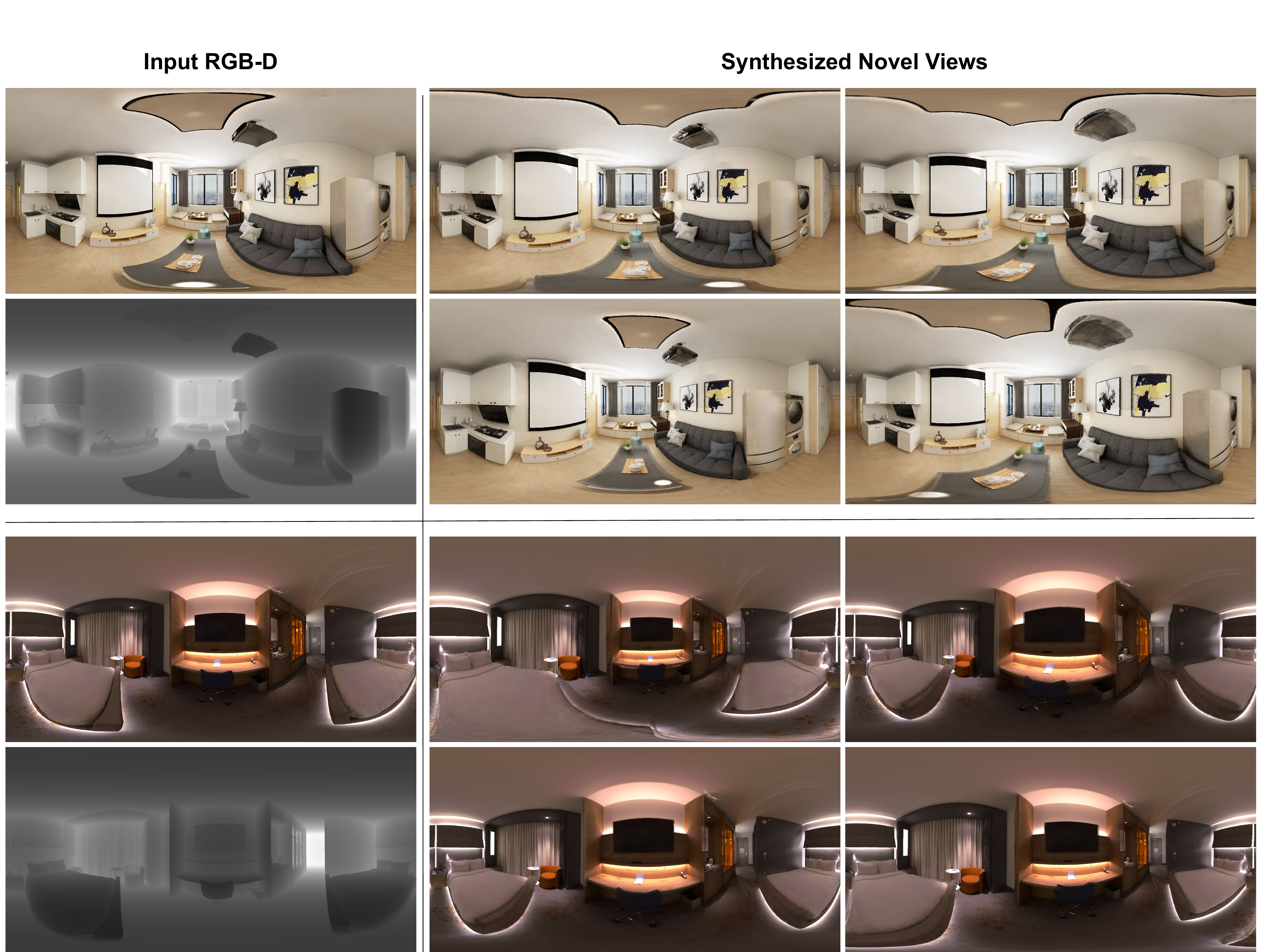}
  \caption{Visualization of novel view rendering in the proposed method on two samples of Structured3D and Replica360. 
  A plausible viewpoint image with 3D consistency is synthesized at positions different from the camera position of the input.}
  \label{fig:visualise}
\end{figure*}

Omnidirectional cameras have become more easily accessible, with a growing number of panoramas shared on media and $360^\circ$ datasets released. $360^\circ$ cameras can capture complete environments in a single shot, which makes $360^\circ$ imagery alluring in many computer vision tasks. They are becoming increasingly popular and widespread in the computer vision community. 
The omnidirectional $360^\circ$ field-of-view captured by these devices is appealing for tasks such as robust, omnidirectional SLAM \cite{won2020omnislam,sumikura2019openvslam}, scene understanding and layout estimation \cite{jin2020geometric, sun2021hohonet,wang2021led2,zeng2020joint}, or VR photography and video \cite{omniphotos,serrano2019motion}. 
While many techniques are proposed to synthesize novel views by taking the perspective image(s) as the input, prior work rarely considers the panorama image as a single source for modeling and rendering. Although perspective images can be acquired conveniently, in order to construct a full scene, it requires a set of dense samples. Furthermore, additional camera variables are essential for estimating relative poses and matching \cite{hsu2021moving}.

Synthesizing novel views with parallax provides immersive 3D experiences \cite{shum2005virtual}. Traditional computer vision solutions employ reconstruction techniques (e.g., structure from motion \cite{hartley2004camera} and image-based rendering \cite{shum2007rendering,shum1999rendering}) using a set of densely captured images. However, these approaches suffer from the cost of matching and reconstruction computation for both time and capacity. The recent development in this field focuses on deep learning methods for its strong capability of modeling 3D geometry and rendering new frames \cite{hsu2021moving}. In recent years, neural network-based rendering methods have been rapidly developed, and the neural radiance field (NeRF) \cite{mildenhall2021nerf} is a promising method for synthesizing photorealistic views. However, the NeRF requires tens to hundreds of images with known relative positions and the same shooting conditions to be given as input, and such imaging is a large and time-consuming process \cite{hara2022enhancement}. Accordingly, various efforts have been made to reduce the number of input images \cite{yu2021pixelnerf,wang2021ibrnet,trevithick2020grf,dietnerf} or ease the shooting conditions \cite{martin2021nerf,wang2021nerf,lin2021barf,jeong2021self}.

We attempt to learn a 3D scene model from a single $360^\circ$ image. Learning NeRF from a single $360^\circ$ image is advantageous because we do not need to align the shooting conditions between images. Furthermore, we do not need to know the relative positions between images because we use only one image that contains a wealth of omnidirectional information \cite{hara2022enhancement}. OmniNeRF \cite{hsu2021moving} is a prior study of this approach; however, it relies only on the neighborhood interpolation capability of the multi-layer perceptron to complete the missing regions caused by occlusion. The single source image does not contain enough information to infer the occlusion and the opposite side of objects \cite{hara2022enhancement}; thus, the results are degraded. \cite{hara2022enhancement} tries to complete the missing regions of the reprojected images by using a self-supervised generative model. However, the model fails when there are large missing areas in the reprojected images.

Only a few NeRF methods have been proposed to take advantage of depth measurements simultaneously with color within the volumetric rendering pipeline \cite{deng2022depth,neff2021donerf}. In this work, we explore depth as an additional, cheap source of supervision to guide the geometry learned by OmniNeRF. We propose to extract the depth information simultaneously when the input image is reprojected at other camera positions. It was noticed that considering depth information can improve geometry considerably compared to only color information.

OmniNeRF is still estimated per scene and cannot benefit from prior knowledge from other images and objects. Prior knowledge is needed when the scene reconstruction problem is underdetermined. 3D reconstruction systems struggle when regions of an object are never observed. This is particularly problematic when rendering an object at significantly different poses. Unobserved regions during training become visible when rendering a scene with an extreme baseline change. A view synthesis system should generate plausible missing details to fill in the gaps. Even a regularized NeRF learns poor extrapolations to unseen regions due to its lack of prior knowledge.

We also exploit the consistency principle utilized in DietNeRF \cite{dietnerf}: objects share high-level semantic properties between their views. Image recognition models learn to extract many such high-level semantic features, including object identity. We transfer prior knowledge from pre-trained image encoders learned on highly diverse 2D single-view image data to the view synthesis problem. In the single-view setting, such encoders are frequently trained on millions of realistic images like ImageNet \cite{deng2009imagenet}. CLIP is a recent multi-modal encoder that is trained to match images with captions in a massive web scrape containing 400M images \cite{radford2021learning}. Due to the diversity of its data, CLIP showed promising zero- and few-shot transfer performance to image recognition tasks. CLIP and ImageNet models also contain prior knowledge useful for novel view synthesis.

Our contributions in this paper are as follows:
\begin{itemize}
    \item We propose 360FusionNeRF, a neural scene representation framework based on OmniNeRF that can be estimated from only a single RGB-D $360^\circ$ Panoramic Image and can generate views with unobserved regions.
    \item In addition to minimizing NeRF’s mean squared error losses at known poses in pixel-space, 360FusionNeRF penalizes a geometric loss via auxiallary depth of the projected images and a self supervised semantic consistency loss via activations of CLIP’s Vision Transformer.
    \item We demonstrate qualitatively and quantitatively that our proposed method results in a generalizable scene representation and improves perceptual quality.
\end{itemize}

\section{Related Work}

\subsection{Neural 3D Rendering}

Neural Radiance Fields (NeRFs) \cite{mildenhall2021nerf} have demonstrated encouraging progress for view synthesis by learning an implicit neural scene representation. Since its origin, tremendous efforts have been made to improve its quality \cite{verbin2021ref,guo2022nerfren,suhail2022light,chen2022aug}, speed \cite{muller2022instant,sun2022direct,fridovich2022plenoxels}, artistic effects \cite{wang2022clip,fan2022unified,jain2022zero}, and generalization ability  \cite{wang2021ibrnet,liu2022neural}. Specifically, Mip-NeRF \cite{barron2021mip} propose to cast a conical frustum instead of a single ray for anti-aliasing. Mip-NeRF 360 \cite{barron2022mip} further extends it to the unbounded scenes with efficient parameterization. KiloNeRF \cite{reiser2021kilonerf} speeds up NeRF by adopting thousands of tiny MLPs. MVSNeRF \cite{chen2021mvsnerf} extracts a 3D cost volume and renders high-quality images from novel viewpoints on unseen scenes. DS-NeRF \cite{deng2022depth} adopts additional depth supervision to improve the reconstruction quality. RegNeRF [34] proposes a normalizing flow and depth smoothness regularization. DietNeRF \cite{dietnerf} utilizes the CLIP embeddings to add semantic constraints for unseen views. PixelNeRF \cite{yu2021pixelnerf} utilizes a ConvNets encoder to extract context information by large-scale pre-training and successfully renders novel views from a single input. However, it can only work on simple objects (e.g., ShapeNet ) \cite{xu2022sinnerf}, while the results on complex scenes remain unknown. Furthermore, the approach relies on the availability of the entire reference image for supervision, but the reprojected images are incomplete in a panoramic setting. SinNeRF \cite{xu2022sinnerf} too proposes a multi-supervision NeRF, but its approaches are very object centric. In our work, we focus on complex scene reconstruction with panoramic images.

\subsection{360 Panorama View Synthesis}

OmniNeRF\cite{gu2022omni} synthesizes novel fish-eye
projection images, using spherical sampling to improve the quality of results. 360Roam\cite{huang2022360roam} is a scene-level NeRF system that can synthesize images of large-scale indoor scenes in real-time and support VR roaming. PanoHDR-NeRF \cite{gera2022casual} presents a pipeline to predict the full HDR radiance of an indoor scene without using special hardware, careful scanning of the scene, or intricately
calibrated camera configurations. However, this paper focuses on synthesizing novel views from a single Equirectangular Panorama $360^\circ$ RGB-D Image.

OmniNeRF \cite{hsu2021moving} learns an entire scene from a single $360^\circ$ RGBD image without the need to set relative positions or identify shooting conditions. However, it only relies on the neighborhood interpolation capability of the multi-layer perceptron to complete the missing regions caused by occlusion and zooming, which leads to artifacts, and the image quality is greatly reduced when moving away from the camera position of the input image \cite{hara2022enhancement}. An alternative method to NeRF, Pathdreamer \cite{koh2021pathdreamer}, synthesizes novel views from a single $360^\circ$ RGB-D image. However, it has the issue of low 3D consistency in the synthesized views due to its reliance on 2D image-to-image translation.

In \cite{hara2022enhancement}, a self-supervised trained generative model completes the missing regions of the reprojected images of OmniNeRF, and the completed images are utilized for training the NeRF. They introduce a method to train NeRF while dynamically selecting a sparse set of completed images, to reduce the discrimination error of the synthesized views with real images. However, when there are large missing regions that exceed the image completion capabilities, it fails to synthesize plausible views.


\section{Proposed Method}

\subsection{Preliminaries}
Neural Radiance Fields (NeRFs) \cite{mildenhall2021nerf} synthesize images sampling 5D coordinates (location $(x, y, z)$ and viewing direction $(\theta, \phi)$) along camera rays, map them to color $(r, g, b)$ and volume density $\sigma$. \cite{mildenhall2021nerf} first propose using coordinate-based multi-layer perception networks (MLPs) to parameterize this function and then use volumetric rendering techniques to alpha composite the values at each location to obtain the final rendered images.

OmniNeRF \cite{hsu2021moving} generates multiple images at virtual camera positions from a single $360^\circ$ RGB-D image and utilizes these images to train NeRF. A set of 3D points is generated from the given RGB-D panorama, and then these 3D points are reprojected into multiple omnidirectional images that correspond to different virtual camera locations. The generated omnidirectional images are likely to be imperfect as there might be gaps and cracks between pixels due to occlusion or limited resolution. OmniNeRF solves this problem by taking advantage of the pixel-based prediction property of its MLP model, which takes a single pixel rather than an entire image as the input.

At each discrete sample on the ray $r(t) = o + td$, where $o$ and $d$ denote the ray origin and ray direction, the final RGB values $C(r)$ are optimized from aggregation of color $c_i$ and opacity $\sigma_i$. A positional encoding technique~\cite{tancik2020fourier} is applied to rays for capturing high-frequency information. The function of color composition follows the rule in volume rendering~\cite{max1995optical}:
\begin{equation}
   \hat{C}(r) = \sum_{i=1}^N T_i \left(1-\exp(- \sigma_i \delta_i)\right) c_i,  \\
\end{equation}
where  $T_i = \exp \left( - \sum_{j=1}^{i-1}\sigma_j \delta_j \right)$ 
and $\delta_i = t_{i+1} - t_i$ is the interval between two adjacent samples. The overall volume sampling principles are done hierarchically: a `coarse' and a `refined' stage. The coarse and refined networks are identical except for the process of sampling pixels on a ray. At the coarse stage, $N_c$ intervals are uniformly sampled alone the ray, while at the refined stage, $N_f$ intervals are decided in accordance with densities from the coarse stage. These two predictions would be optimized by the ground-truth color, respectively. It optimizes the radiance field by minimizing the mean squared error between rendered color and the ground truth color,
\begin{equation}
\label{eqn:color-loss}
    \mathcal{L}_{\text{Color}} = \sum_{r \in R_{i}}|| (C(r) - \hat{C}(r)) ||^2
\end{equation}
where $R_i$ is the set of input rays during training.

\begin{figure*}[ht!]
    \centering
  \includegraphics[width=0.99\linewidth]{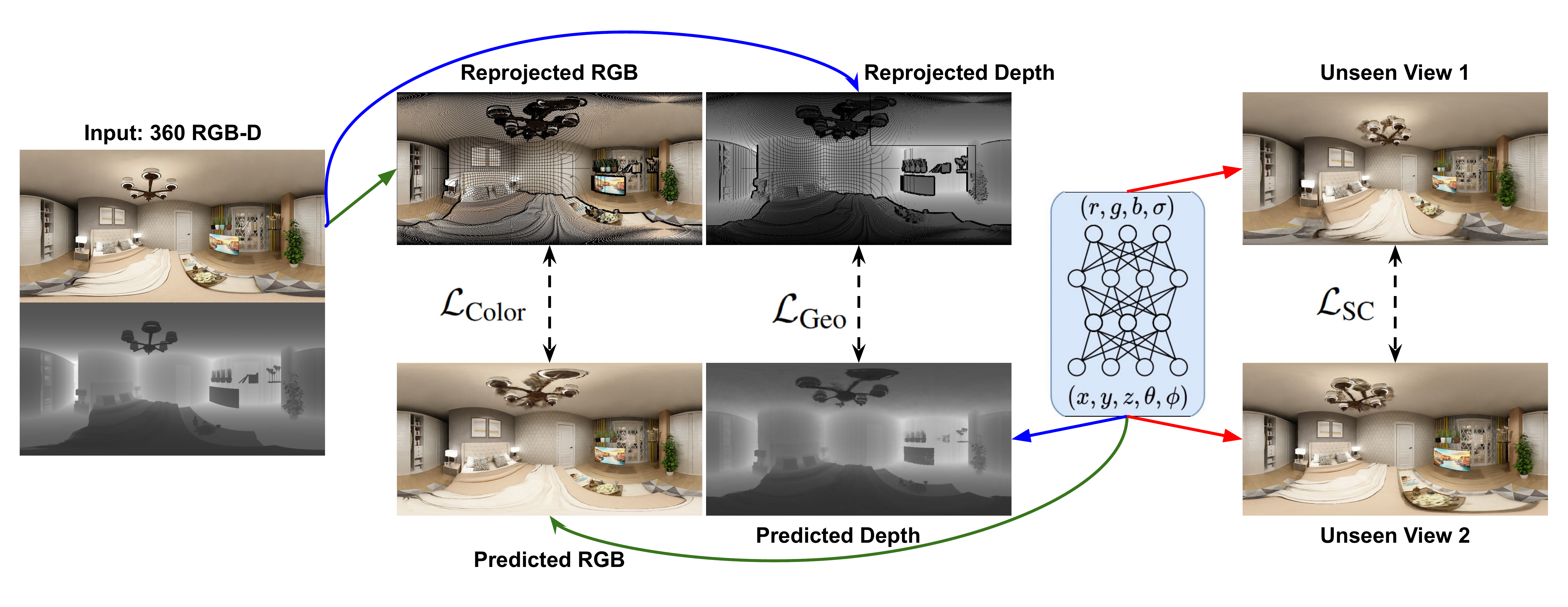}
  \caption{The input data includes a panorama and an auxiliary depth map. We generate new training images, and depth maps with various virtual camera poses. Since the information might be missing after re-projection, we could only have partial pixels in each augmented training image. The Projected RGB Images are utilized for Color Supervision, while the Depth maps are used for Geometric Supervision. The Novel Views contribute toward maintaining Semantic Consistency using CLIP ViT. }
  \label{fig:method}
\end{figure*}

\subsection{Challenges with OmniNeRF}

\subsubsection{Overfitting to Training Views}
Conceptually, OmniNeRF is trained by mimicking the image-formation process at observed poses. With many training views, the MLP in OmniNeRF recovers accurate textures and occupancy that allow interpolations to new views. The high-frequency representational capacity allows OmniNeRF to overfit to each input view. Fundamentally, the plenoptic function representation suffers from a near-field ambiguity \cite{zhang2020nerf++} where distant cameras each observe significant regions of space that no other camera observes. In this case, the optimal scene representation is underdetermined. Degenerate solutions can also exploit the view dependence of the radiance field. \cite{dietnerf} had pointed out that while a rendered view from a pose near a training image has reasonable textures, it is skewed incorrectly and has cloudy artifacts from incorrect geometry. As the geometry is not estimated correctly, a distant view contains almost none of the correct information. High-opacity regions block the camera. Without supervision from any nearby camera, opacity is sensitive to random initialization.

\subsubsection{No Generalization to Unseen Views}
As OmniNeRF is estimated from scratch per scene, it has no prior knowledge about natural objects such as common symmetries and object parts. The fundamental challenge is that NeRF receives no supervisory signal from $\mathcal{L}_{\text{Color}}$ to the unobserved regions and instead relies on the inductive bias of the MLP for any inpainting. We want to introduce prior knowledge that allows NeRF to exploit bilateral symmetry for plausible completions.

\subsection{Geometric Supervision}
Directly overfitting the reference images leads to a corrupted neural radiance
field collapsing towards the provided views. We start by adopting the depth prior to reconstructing reasonable 3D geometry.
\subsubsection{Extracting Depth Ground Truths}
We follow the projection procedure from OmniNeRF \cite{hsu2021moving}] to extract depth information simultaneously with the view. First, all pixels can be projected to a uniform sphere by their 2D coordinates. For a pixel $(x, y)$ on the panorama, its vertical and horizontal viewing angles can be defined by $\theta = \pi y / H $, $\phi = 2 \pi x / W $, where $H$ and $W$ are the height and width of the panorama. The coordinate center would be the current camera position, namely the ray origin. Likewise, a ray direction means a unit vector from the center to the sphere. Therefore, a novel panoramic view and the depth information can be determined by moving the camera to a new position and examining what would be sampled on the new sphere by the emitted rays based on the above equations. Not all pixels are supposed to be visible from the new viewpoint. The key of the projection mechanism is to verify which parts of the ground truth will be visible to a given ray origin.

\subsubsection{Volumetric Rendering}

Similar to Color rendering, the depth can be represented with
volume density using:
\begin{equation}
   \hat{D}(r) = \sum_{i=1}^N T_i \left(1-\exp(- \sigma_i \delta_i)\right) t_i,  \\
\end{equation}
where  $T_i = \exp \left( - \sum_{j=1}^{i-1}\sigma_j \delta_j \right)$ 
and $\delta_i = t_{i+1} - t_i$ is the interval between two adjacent samples.

\subsubsection{Optimization}

The network parameters $\theta$ are optimized using a set of RGB-D frames, each of which has a color, depth, and camera pose information. $\mathcal{L}_{\text{Color}}$ in Equation \ref{eqn:color-loss} acts as the photometric loss. The geometric loss is the absolute difference between predicted and true depths, normalized by the depth variance \cite{sucar2021imap} to discourage weights with high uncertainty: the geometric loss is given by:

\begin{equation}
    \mathcal{L}_{\text{Geo}} = \sum_{r \in R}\frac{\lvert \hat{D}(r)-D(r) \rvert}{\sqrt{\hat{D}_{var}(r)}},
\end{equation} where \(\hat{D}_{var}(r) = \sum_{i=1}^{N} T_i(1 - exp(-\sigma_i\delta_i))(\hat{D}(r) - t_i)^2\) depth variance of the image.

\subsection{Semantic Consistency}

Unlike the geometry pseudo labels, where we enforce the consistency in 3D space,
pseudo semantic labels are adopted to regularize the 2D image fidelity. Concretely speaking, we introduce a global structure prior supported by a pre-trained
ViT network. This guidance helps SinNeRF render visually-pleasing results in each view.

Vision transformers (ViT) have been proven to be an expressive semantic prior, even between images with misalignment \cite{tumanyan2022splicing} \cite{amir2021deep}. Similar to \cite{xu2022sinnerf}, we propose to adopt a pre-trained ViT for global structure guidance, which enforces semantic consistency between unseen views. Although pixel-wise misalignment exists between the views, we agree with the observation by \cite{xu2022sinnerf} that the extracted representation of ViT is robust to this misalignment and provides supervision at the semantic level. Intuitively, this is because the content and style of the two views are similar, and a deep network is capable of learning invariant representation.

Here we adopt CLIP-ViT \cite{radford2021learning}, a self-supervised vision transformer trained on ImageNet \cite{deng2009imagenet} dataset. In practice, CLIP produces normalized image embeddings. When the embedding is a unit vector, the $\mathcal{L}_{\text{SC}}$ simplifies to cosine similarity up to a constant and a scaling factor that can be absorbed into the loss weight $\lambda$:

\begin{equation}
    \mathcal{L}_{\text{SC}}(I_1, I_2) = \lambda \phi(I_1)^T \phi(I_2),
\end{equation}

where $\phi$(.) is the normalized image embedding and $I_1$, $I_2$ are unseen views.

\subsection{Final Pipeline}

For generating $\mathcal{L}_{\text{SC}}$, volume rendering is necessary, and it is computationally expensive. Hence, semantic consistency is computed over a smaller resolution of the views. Further, as observed by \cite{dietnerf}, $\mathcal{L}_{\text{SC}}$ converges faster as compared to  $\mathcal{L}_{\text{Color}}$ and $\mathcal{L}_{\text{Geo}}$. Hence, $\mathcal{L}_{\text{SC}}$ is minimized every $k$ iterations for every minimization of $\mathcal{L}_{\text{Color}}$ + $\lambda_{\text{Geo}} \mathcal{L}_{\text{Geo}}$.

\section{Experiments}

\begin{figure*}[!ht]
    \centering
  \includegraphics[width=0.99\linewidth]{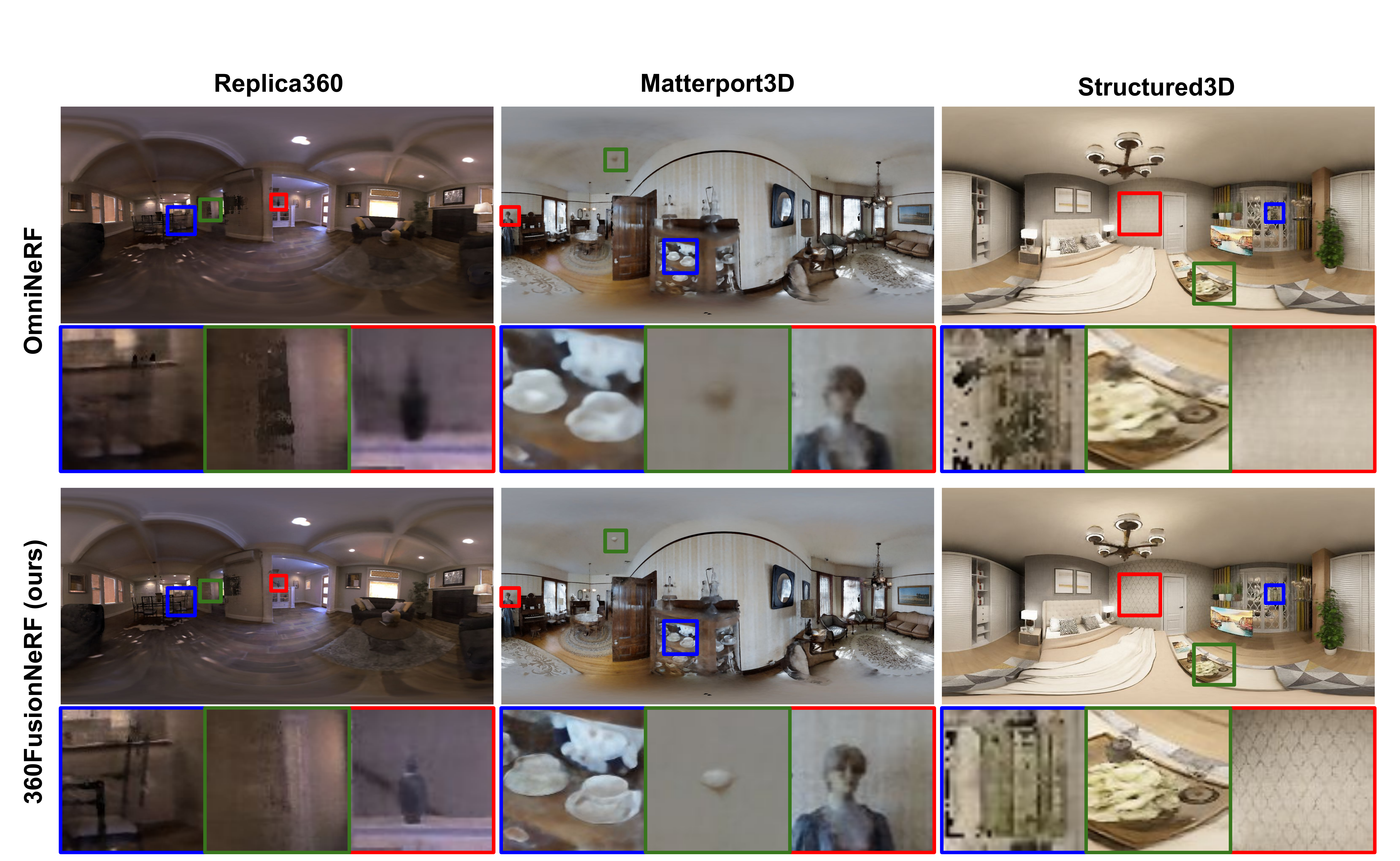}
  \caption{Qualitative comparison of OmniNeRF and the proposed method on Replica360, Matterport3D, and Structured3D datasets. }
  \label{fig:qualitative}
\end{figure*}

\begin{table*}
\setlength{\tabcolsep}{8.5pt}
\center
\begin{tabular}{@{}l |  c c c | c c c | c c c}
\toprule
 & PSNR$\uparrow$ & SSIM$\uparrow$ & LPIPS$\downarrow$ & PSNR$\uparrow$ & SSIM$\uparrow$ & LPIPS$\downarrow$ & PSNR$\uparrow$ & SSIM$\uparrow$ & LPIPS$\downarrow$\\
\midrule

Structured3D & \multicolumn{3}{c}{S1} & \multicolumn{3}{c}{S2} & \multicolumn{3}{c}{S3}  \\
\midrule

OmniNeRF & 26.41 & 0.8259 & 0.2860 & 22.29 & 0.8627 & 0.2614 & 29.72 & 0.8903 & 0.2472  \\ 
360FusionNeRF (ours) & \textbf{28.05} & \textbf{0.8734} & \textbf{0.2260} & \textbf{23.45} & \textbf{0.8731} & \textbf{0.2570} & \textbf{30.20} & \textbf{0.9061} & \textbf{0.2047} \\
\midrule
Matterport3D & \multicolumn{3}{c}{M1} & \multicolumn{3}{c}{M2} & \multicolumn{3}{c}{M3}  \\
\midrule

OmniNeRF & 25.01 & 0.8860 & 0.2720 & \textbf{25.61} & 0.8000 & 0.2994 & 19.01 & 0.8481 & 0.2948  \\ 
360FusionNeRF (ours) & \textbf{26.88} & \textbf{0.8934} & \textbf{0.2573} & 25.53  & \textbf{0.8336} & \textbf{0.2575} & \textbf{19.13} & \textbf{0.8622} & \textbf{0.2748} \\

\midrule
Replica360 & \multicolumn{3}{c}{R1} & \multicolumn{3}{c}{R2} & \multicolumn{3}{c}{R3}  \\
\midrule

OmniNeRF & 28.83 & 0.9226 & 0.2715 & 30.61 & 0.9374 & 0.3385 & 27.39 & 0.8865 &  0.3701 \\ 
360FusionNeRF (ours) & \textbf{32.76} & \textbf{0.9451} & \textbf{0.2116} & \textbf{33.23} & \textbf{0.9520} & \textbf{0.2790}  & \textbf{27.61}  & \textbf{0.8951} & \textbf{0.3184} \\



\bottomrule
\end{tabular}
\caption{ Quantitative evaluation of each novel view synthesis method on 3 scenes for each dataset.}
\label{tab:quantitative}
\vspace{-1em}
\end{table*}

\subsection{Dataset}

We test our method on both synthetic and real-world
datasets. In this work, all the panorama images are under
equirectangular projection at the resolution of 512 × 1024 for Structured3D dataset, 1024 $\times$ 2048 for Replica360 and Matterport3D datasets. We randomly select 3 scenes from each dataset and perform a quantitative and qualitative analysis on the same.

\subsubsection{Structured3D}
Structured3D dataset \cite{zheng2020structured3d} contains 3,500 synthetic departments (scenes) with 185,985 photorealistic panoramic renderings. As the original virtual environment is not publicly accessible, we utilized the rendered panoramas directly. 

\subsubsection{Matterport3D}
Matterport3D dataset \cite{chang2017matterport3d} is a large-scale indoor real-world $360^\circ$ dataset, captured by Matterport’s Pro 3D camera in 90 furnished houses (scenes) . The dataset provides 10,800 RGB-D panorama images, where we find the RGB-D signals near the polar region are missing.

\subsubsection{Replica360}
Replica360 dataset \cite{straub2019replica} contains 18 highly photo-realistic 3D indoor scene reconstructions at room and building scale.

\subsection{Implementation Details}
\label{subsec:impl_details}

The Adam optimizer \cite{kingma2015ba} is used for the overall training process. The learning rate is initialized to $5.10^{-4}$, which is then exponentially reduced to $5.10^{-5}$. 200,000 epochs train the model for each experiment with a batch size of 1,400 on a DGX A100 GPU. We set $N_c$ = 64 and $N_f$ = 128 in the coarse and refined networks. The network architecture is identical to that of OmniNeRF \cite{hsu2021moving}.

\subsection{Qualitative Evaluation}

We qualitatively validate the novel view synthesis using a single $360^\circ$ RGB-D image. Figure \ref{fig:qualitative} compares the synthesized novel views by the proposed method and OmniNeRF. Each column corresponds to a scene of a dataset, and each row contains the results of a method. One can see that our method preserves the best geometry as well as perceptual quality. 

In the Replica360 sample, we can see the vase has been blurred for predictions by OmniNeRF while its shape has been well restored in our method. Some artifacts are created on the walls while they have been much reduced in that case. Even though the chair is thin, our method has produced the shape well.

Matterport3D dataset has panoramas blurred at the poles. This makes it difficult for OmniNeRF to reproduce an object near the ceilings or the floor. As seen, the ceiling light has blurred in the background, while it can still be identified by our method. Even the face of the idol and the cups are much clearer and have not lost shape compared to OmniNeRF.

In view synthesis for the Structured3D sample, the texture of the walls has been lost in the case of OmniNeRF while it is well maintained by our method. The artifacts are prevented near the bookshelf, and even transparent objects like the glass cup have not collapsed as in the case of OmniNeRF.

\subsection{Quantitative Evaluation}

We quantitatively evaluated each method using the following three evaluation metrics:
\subsubsection{PSNR}
Peak-to-signal-noise ratio expresses the mean-squared error in log space. This metric evaluates the performance of the input image reconstruction. We calculate
the PSNR between the reference image and the synthesized image at the position of the input image.

\subsubsection{SSIM}
Structural Similarity Index Measure \cite{wang2004image} quantifies the degradation of image quality in the reconstructed image; the higher is better. However, it often disagrees with human judgements of similarity \cite{zhang2018unreasonable}.

\subsubsection{LPIPS}
Deep CNN activations mirror aspects of human perception. We measure the perceptual image quality using LPIPS \cite{zhang2018unreasonable}, which computes MSE between normalized features from all layers of a pre-trained VGG encoder \cite{simonyan2015very}.

We extract three scenes each from the Structure3D, Matterport3D and Replica360 datasets. Table \ref{tab:quantitative} presents the evaluation results for each image. In almost all scenes, the proposed method outperforms OmniNeRF in terms of the PSNR, SSIM and LPIPS, which indicates that the proposed method can better synthesize plausible views with features close to the dataset. 

The lower performance for both models in Matterport3D can be due to distortion caused by the blurring of the panoramic images at the poles. In M2, the PSNR score is better for OmniNeRF, while our method outperforms both SSIM and LPIPS. Because of uncertainty, blurry renderings will outperform sharp but incorrect renderings on average error metrics like MSE and PSNR. Arguably, perceptual quality and sharpness are better metrics than pixel error for graphics applications like photo editing and virtual reality, as plausibility is emphasized.

\section{Conclusions}

This paper proposes a method for synthesizing novel views by learning the neural radiance field from a single $360^\circ$ image. The proposed method reprojects the input image to $360^\circ$ images at other camera positions, and its depth map is estimated. A geometric loss and semantic consistency loss were introduced in addition to the Color loss. Experiments indicated that the proposed method could synthesize plausible novel views while preserving the features of the scene for artificial and real-world scenes. These results confirm the effectiveness of employing geometric and semantic supervision for panoramic novel view synthesis.

\section{Acknowledgment}
This research was supported by grants from NVIDIA and utilized NVIDIA SDKs (CUDA Toolkit, TensorRT, and Omniverse).

\bibliographystyle{IEEEtran}
\bibliography{IEEEexample}
\end{document}